\documentclass[preprint,review,12pt]{elsarticle}

\usepackage[margin=2.5cm]{geometry}
\usepackage{tabu}
\usepackage{colortbl}
\usepackage{multicol}
\usepackage[font=small,skip=0pt]{caption}
\usepackage{subcaption}
\usepackage{multirow}
\usepackage{graphicx}
\usepackage[skins]{tcolorbox}
\usepackage{tikz}
\usetikzlibrary{calc,matrix,shapes,arrows,positioning,decorations.pathreplacing}
\usetikzlibrary{mindmap,backgrounds}
\usetikzlibrary{calc,fit,arrows}
\usetikzlibrary{intersections}
\usepackage{algorithm}
\usepackage{algpseudocode}% http://ctan.org/pkg/algorithmicx
\usepackage{adjustbox}
\usepackage{collcell,datatool}
\usepackage{array}
\usepackage{mathtools}
\usepackage{cancel}
\usepackage{amsthm,amssymb,amsmath}
\usepackage{framed} 
\usepackage{lastpage}

\usepackage{makeidx}
\makeindex

  \usepackage{pgfplots, pgfplotstable}
\usepgfplotslibrary{fillbetween}
\usepackage{filecontents}
  \pgfplotsset{compat=newest}
  \usetikzlibrary{plotmarks}
  \usetikzlibrary{arrows.meta}
  \usepgfplotslibrary{patchplots}
  \usepackage{grffile}
  \usepackage{amsmath}
\usetikzlibrary{patterns}
\usetikzlibrary{backgrounds}

\usepackage{enumitem}

\DeclareMathOperator*{\diag}{diag}

\algnewcommand{\LeftComment}[1]{\Statex \(\triangleright\) #1}
\usepackage{pdflscape} 
\usepackage{lineno, hyperref}

\hypersetup{
  colorlinks=false,
  urlbordercolor=white
}

\modulolinenumbers[5]
\journal{arxiv}

\biboptions{authoryear}

\newtheorem{theorem}{Theorem}[section]

\newtheorem{proposition}[theorem]{Proposition}

\newcommand \EquationRef[1]{\eqref{#1}}
\newcommand \SectionRef[1]{Section~\ref{#1}}
\newcommand \AlgoRef[1]{Algorithm~\ref{#1}}
\newcommand \TableRef[1]{Table~\ref{#1}}
\newcommand \FigRef[1]{Fig.~\ref{#1}}
\newcommand \FigPath[1]{./Picture/#1}

\begin{document}

\begin{frontmatter}
	    
    \title{FRMDN: Flow-based Recurrent Mixture Density Network}
    \author[ut]{Seyedeh Fatemeh Razavi}
    \ead{razavi\_f@ut.ac.ir}
    \author[ut]{Reshad Hosseini}
	\ead{reshadh@ut.ac.ir}
	\author[ut]{Tina Behzad}
	\address[ut]{School of ECE, College of Engineering, University of Tehran, Tehran, Iran}
	\ead{tina.behzad@ut.ac.ir}
    
    \begin{abstract}
        The class of recurrent mixture density networks is an important class of probabilistic models used extensively in sequence modeling and sequence-to-sequence mapping applications. In this class of models, the density of a target sequence in each time-step is modeled by a Gaussian mixture model with the parameters given by a recurrent neural network. In this paper, we generalize recurrent mixture density networks by defining a Gaussian mixture model on a non-linearly transformed target sequence in each time-step. The non-linearly transformed space is created by normalizing flow. We observed that this model significantly improves the fit to image sequences measured by the log-likelihood. We also applied the proposed model on some speech and image data, and observed that the model has significant modeling power outperforming other state-of-the-art methods in terms of the log-likelihood. 
    \end{abstract}
    

\end{frontmatter}

%\linenumbers

\section{Introduction}
\label{sec:introduction}
% \input{Sections/Introduction}
% Definition of Sequential Conditional Model
Sequential conditional density models are used in sequential data generation, prediction of time series, and sequential decision making \citep{Graves2013,WaveGlow,WorldModelNips,Seq2SeqNips,VariationalRNN}.
In these models, the target sequence in a time-step is modeled based on the target sequence in previous time-steps and the input sequence.
In these models, there are two random variables \(\mathcal{X}\) and \(\mathcal{Y}\) called conditional and dependent sequential variables, respectively.

% Difference with regression
In conditional density models, the full probability distribution function \(p(y|x)\) is modeled, while a regression model only computes a point estimate which usually minimizes a mean-squared loss given by \(\mathbb{E}[y|x]\). The point estimate given by the regression model is not a good estimate when we have a multi-modal distribution for some \(y\) value. In these situations, it is common to use conditional density modeling and calculate a point estimate from it. An important choice is using a Gaussian Mixture Model (GMM) with the parameters be functions of the conditional random variable.

% Mixture Density Network
An expressive framework for modeling a conditional distribution is using a Mixture Density Network (MDN) \citep{MDN1994,BishopBook2006}, which is a model that uses a neural network to parameterize a GMM. In other words, an MDN puts a mixture distribution, specifically a GMM in the last layer of a neural network.
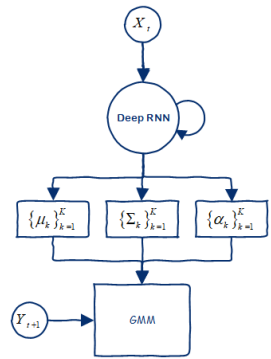
Based on applications, different kinds of neural networks such as feed-forward \citep{MDN1994,BishopBook2006}, recurrent \citep{Graves2013,BidirectionalRMDN}, convolutional \citep{ConvolutionalMDN}, fuzzy \citep{fuzzy}, and graph \citep{GraphMDN} neural networks have been used in the literature. A sequential version of MDNs called Recurrent Mixture Density Network (RMDN) have been successfully used in many applications such as handwritten synthesis \citep{Graves2013}, sketch drawing \citep{Sketch}, reinforcement learning \citep{WorldModelNips}, spatiotemporal visual attention \citep{VisualAttention}, demand forecasting \citep{TaxiDemand}, trajectory prediction \citep{Trajectory,Limitation,Autonomous,PredictLocation,Primitive}, recommendation systems \citep{AttentionRMDN}, acoustic modeling in speech synthesis \citep{Speech2014,AcousticModel,AR_Speech}, uncertainty estimation \citep{Uncertainty,Uncertatinty1} and so on.

% Recurrent Mixture Density Network
RMDN uses a Recurrent Neural Network (RNN) to generate the parameters of a GMM in every time-step. Although GMM is a flexible distribution used in each time-step of RMDN, it works best for modeling densities that are clustered in the space.
Sequential data do not necessarily satisfy this model, and sometimes datapoints are very cluttered in the space. To overcome this limitation, we propose to use a mapping from a space with a cluttered data to a well-clustered one before using a GMM. Fortunately, Normalizing Flows (NFs) have been introduced recently and match our need for having a closed-formed likelihood function \citep{SurveyNF}. These networks try to transfer source data with a complex distribution to a much simpler target distribution with an invertible transformation function.

% Limitation 1
Recently, it has been shown that combining RNNs and NFs shows advantages in density modeling.
A group of researchers use a RNN to parameterize a NF \citep{TransformerFlow}.
\citep{RFN_PR} uses a NF instead of a Gaussian in the decoder part of a variational RNN.
Another group assumed different dimensions of input data as time-steps, and used NFs with uni-dimensional RMDNs to model the data \citep{TAN}.
This idea is close to this work, but it treats high-dimensional data as a sequence of single-variable data and does not propose any solution for high-dimensional sequential data.
Another related work is \citep{GaussianCopula}, where the authors use a point-wise nonlinear transformation in a RMDN.

% Limitation 2
Moreover, commonly used RDMNs make another assumption of having diagonal covariance matrices for reducing the number of parameters and making the optimization problem easier
\citep{MDN1994, Speech2014, AR_Speech, Uncertainty, RFN_PR}.
There are works that use full covariance matrices for Gaussian densities and use Cholesky decomposition for solving the problem of having a positive definite constraint of the covariance matrices \citep{Choleski,FaceAlignment,RegMDN,NNforConditional96}.
These works still suffer from the explosion of the parameters for high-dimensional data, where many components are used for modeling.
Considering different kinds of decompositions for the covariance matrix to reduce the number of parameters has been investigated by researchers.
One of the works shares parts of decomposed parameters among components of a GMM to achieve efficient high-dimensional computation \citep{TiedGMM}.
Another work considers the sum of a diagonal and a low-rank matrix as the covariance matrix \citep{GaussianCopula}.

% Proposed Method
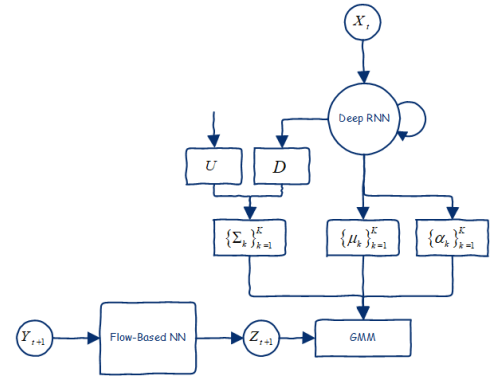
In this paper, we introduce a Flow-based Recurrent Mixture Density Network (FRMDN) to overcome the limitations of RMDN.
Evaluation of the proposed method has been done for three applications of image sequence, speech and image modeling.
Our main contributions in this paper are as follows.
More details about the proposed method can be found in Section \ref{sec:proposed}.

% Our contributions
\begin{itemize}

	\item
	We use NF to transform the dependent variable in each time-step, so to increase the flexibility of RMDNs for the data that are not well-clustered in the space. Since NFs are invertible and have a tractable way to compute the logarithm of the determinant, we can minimize Negative Log-Likelihood (NLL) directly and can sample from the underlying conditional sequential distribution. Mathematically speaking, a GMM is applied on \(f(y_t)\) where \(f\) is an invertible mapping like NF.
	
	\item
	Moreover, we consider a decomposition for precision matrices (inverse of the covariance matrices) defined as \(\Sigma _k^{ - 1} = D_k+U_kU_k^T\) for every component in the mixture distribution for both RMDN and FRMDN.
	Based on this decomposition, every component in GMM has a specific diagonal matrix \(D_k\), and a low-rank matrix constructed by a \(d \times d\) non-square matrix \(U_k\), for \(d'\ll d\). We will show that we can have efficient computations for FRDMN using this decomposition.
	
\end{itemize}

% Experiments 1
At first, applicability of the proposed method is examined for 
modeling image sequences. Modeling image sequences has been used in
a well-known model-based Reinforcement Learning (RL) problem, called the world model \citep{WorldModelNips}.
Using similar logic as in \citep{WorldModelNips}, a vision unit is used with a convolutional variational auto-encoder to encode each image in a compact representation.
Then, FRMDN instead of RMDN is used to model the encoded image sequence. The experimental results verify the improvement of the proposed method in comparison with a basic RMDN in terms of the NLL.

% Experiments 2
The second experiment is done for speech modeling on three datasets (Blizzard, TIMIT, Accent). The experiments indicate the superiority of the proposed method in terms of the NLL.
It should be mentioned that in this experiment, we report the results for diagonal covariance matrices for FRMDNs, because adding a low-rank matrix to a diagonal matrix does not improve the results.

% Experiments 3
In end, the applicability of the proposed method is examined for image modeling on two datasets (MNIST, CIFAR10).
The fairly good quality of the generated images on MNIST dataset and defeating state-of-the-art auto-regressive methods in terms of the NLL verify the observed performance in the two previous applications.

% Organization of paper
The paper is organized as follows.
\SectionRef{sec:background} reviews some related works to this research.
\SectionRef{sec:preliminaries} briefly introduces the preliminaries of this research, such as RMDN and NFs.
\SectionRef{sec:proposed} proposes FRMDN.
Experimental results of FRMDN for three applications, sequence image modeling, speech modeling, and single image modeling, are presented in  \SectionRef{sec:experiment}.
Finally, the paper is concluded in \SectionRef{sec:conclusion}.

\section{Background}
\label{sec:background}
Sequential conditional density models can be used as generative models, where we only have one sequence and the model predicts the value in a time-step from its previous values.
We first give a short review of the literature of sequential generative models.
At the end of this section, an overview of the limitations of RMDNs is presented.

\subsection{Sequential generative models}
\label{sec:sgm}

We focus on commonly used deep sequential generative models.
A diagram of these methods is shown in \FigRef{fig:diagram_sgm}. Our proposed methods is indicated as green in this diagram.

\begin{figure}
\centering
\begin{adjustbox}{width=\textwidth}
	\begin{tikzpicture}
%[sibling distance=100mm, level distance=10mm,
%[node distance=20cm,
[
level 1/.style={sibling distance=200mm}, %, sibling angle=60},
level 2/.style={sibling distance=100mm},
level 3/.style={sibling distance=40mm},
level 4/.style={sibling distance=40mm},
level 5/.style={sibling distance=50mm},
level 6/.style={sibling distance=50mm},
every node/.style = {minimum size=1cm, font=\large \bfseries, shape=rectangle, rounded corners,
	draw, align=center,
	top color=white, bottom color=blue!20}]
\node [minimum size=1cm, text width=10cm]{\Large \bf Sequential generative models}
% non-sequential ---------------------------------
%[{sibling distance=5mm}]
% child
% %[sibling angle=180]
% 	{
% 		node [minimum size=1cm]{\large \bf non-sequential}
% 		% non-sequential and likelihood-based
% 		child {node [minimum size=1cm]{\large \bf likelihood-based}
% 				% non-sequential and likelihood-based and non-conditional
% 				child {node [minimum size=1cm]{\large \bf non-conditional}
% 							child{node {VAE}}
% 							child{node {NF}}
% 							child{node {DM}}
% 						}
% 				% non-sequential and likelihood-based and conditional
% 				child{node {conditional}
% 							child{node {CVAE}}
% 							child{node {CNF}}
% 							child{node {CDM}}
% 						}
% 				}
% 		% non-sequential and non-likelihood-based
% 		child {node [minimum size=1cm]{\large \bf non-likelihood-based}
% 				% non-sequential and likelihood-based and non-conditional
% 				child{node {non-conditional}
% 							child{node {GAN}}
% 						}
% 				% non-sequential and likelihood-based and conditional
% 				child{node {conditional}
% 							child{node {CGAN}}
% 						}
% 				}
% 		}
% sequential ---------------------------------
% child {
% 		node [minimum size=1cm]{\large \bf sequential}
		% non-sequential and likelihood-based
		child {node [minimum size=1cm]{\large \bf likelihood-based}
				% non-sequential and likelihood-based and non-conditional
				child{node {non-conditional}
					child{node {AR}
							child{node [top color=white, bottom color=green!50]{RMDN}}
							}
					child{node {HMM}}
				}
				% non-sequential and likelihood-based and conditional
				child {node {conditional}
					child{node {CHMM}}
					child{node {CAR}
						child{node [top color=white, bottom color=green!50]{CRMDN}}							
						child{node {encoder-decoder}
							child{node {RNN based}}
							child{node {RNN with attention}}
							child{node {transformer}}
						}
					}
					child{node {CTC}}
				}
			}
		% non-sequential and non-likelihood-based
		child {node [minimum size=1cm]{\large \bf non-likelihood-based}
				% non-sequential and likelihood-based and non-conditional
				child{node {non-conditional}
					child{node {GAN}}
				}
				% non-sequential and likelihood-based and conditional
				child{node {conditional}
					child{node {CGAN}}
				}
				};
% 		};
\end{tikzpicture}
\end{adjustbox}
\caption{A hierarchical diagram of deep sequential generative models discussed in \SectionRef{sec:background}. All deep generative models can  be classified according to three indicators (type of objective function, nature of data, and conditionality or non-conditionality of the problem). The full phrases of some abbreviated names on the leaves of this hierarchical diagram from left to right with the removal of duplicates are as follows:
Auto-Regressive (AR), Hidden Markov Model (HMM), Conditional Hidden Markov Model (CHMM), Conditional Recurrent Mixture Density Network (CRMDN), Connectionist Temporal Classification (CTC), and Conditional Auto-Regressive (CAR), Generative Adversarial Network (GAN), Conditional Generative Adversarial Network (CGAN).
}
\captionsetup{justification=centering,margin=2cm}
\label{fig:diagram_sgm}
\end{figure}

% sequential non-conditional likelihood based generative models ==>> AR, HMM
% sequential non-conditional likelihood based generative models ==>> AR
Most famous likelihood-based generative sequential models can be categorized in two main category Auro-Regressive (AR) models and Hidden Markov Models (HMMs). In AR models, the total distribution as a product of conditional distributions of the sequence in each time-step conditioned on previous time-steps. RNNs can be used in AR models, where hidden states in each time-step in this networks accumulate the information of previous time-steps \citep{DeepBook}. Other networks such as convolutions networks can be used to calculate conditional distribution \citep{PixelRNN}. RMDN is a RNN-based model that uses a mixture distribution for modeling the conditional densities. RMDN has a relatively old background \citep{MDN1994,BishopBook2006,BidirectionalRMDN,NNforConditional96}, but it has been revived recently by a research about generating  handwritten sequences \citep{Graves2013}. Afterward, it has been used in other tasks \citep{WorldModelNips,Sketch,VisualAttention,Trajectory,Limitation,Autonomous,PredictLocation,Primitive,AttentionRMDN,Speech2014,AcousticModel,AR_Speech,Uncertainty,Uncertatinty1}.

% sequential non-conditional likelihood based generative models ==>> HMM
HMM is a special case of directed graphical models used for sequential density modeling \citep{BishopBook2006}.
They have two groups of variables named hidden states (which are discrete) and observations (which can be discrete or real).
To avoid ambiguity about the naming of variables in this paper and the literature of HMMs, it is worth noting that the latent variables will be equivalent to conditional variables, while observations will be equivalent to dependent variables.
The probability of hidden states at the beginning time-step, transition probability between hidden states, and emission probability of observation given hidden states at each time-step are the parameters of HMMs.
The probability density of the observation sequence is computed by marginalizing the joint density of hidden states and observations in HMMs.
In HMMs, neural networks can used for computing the conditional probability of each observation given the hidden state. While, early models used GMM for computing this probability. It has been observed that neural network based HMMs commonly outperform GMM-based HMMs \citep{HMM}.

% sequential conditional likelihood based generative models ==>> CAR, HMM, CTC
Sequential generative models are not limited to the above cases and there are a lot of variations in them.
One major group of variations are conditional cases.
The most famous sequential conditional generative models can be categorized as Conditional Auto-Regressive (CAR) models, HMMs, and Connectionist Temporal Classification (CTC) models. In CAR models, a neural networks can be used to compute the density of the output sequence in each time-step given the input sequence and the output sequence of the previous time-steps. The total distribution is the product of this conditional densities. RMDNs can be used for CARs where a RNN is used to parametrize a GMM  \citep{BishopBook2006}. In other words, it puts a mixture distribution, specifically a GMM, in the last layer of a RNN to compute the conditional density.

% sequential conditional likelihood based generative models ==>> CAR (RNN)
An important class of CAR models that worked successfully in many applications is the class of encoder-decoder models. Encoder-decoder models try to encode an input sequence to a context vector, then passes the context vector through another network named decoder to sequentially decode the output sequence. The first architectures of encoder-decoder models used RNNs for both encoder and decoder units \citep{Seq2SeqNips,citekey2}.
However, modeling long sequences is a challenging task with this basic architecture. Because long-term dependencies can be lost in this case and the context vector can not represent the knowledge well enough.

% sequential conditional likelihood based generative models ==>> CAR (attention)
To overcome the problem of long sequences, attention-based models have been proposed in the literature. This approach lets the model focus on different parts of input at each output time-step \citep{Attention1,Attention2}. In this model, based on the decoder network state in
each time-step, a context vector is created from the hidden states of the encoder based on an attention mechanism. 

% sequential conditional likelihood based generative models ==>> CAR (transformer)
It has been observed in the literature that RNN is not necessary in the encoder-decoder model and transformer networks that use simple feed-forward network together with attention mechanism works well \citep{Transformer}. In many applications, such as speech recognition and machine translation, transformers are currently the-state-of-the-art model. Apart from their hight accuracy, they are parallelizable, while the basic framework with RNNs can not be parallelize easily because of its sequential processing of their input.
Although the transformer model has lots of advantages, it is suitable in the case of fixed-length sequences. While RNN-based models are free in terms of sequence length. This problem has been solved in the next generation of transformers named transformer-XL \citep{TransformerXL} by using a recursive module in transformers. 

% sequential conditional likelihood based generative models ==>> HMM
In classical HMMs used in discriminative tasks, the probability of the label given the observation sequence is computed using Bayes rule. 	However in then case when  we have a sequence of target labels, a specific logic is exploited to make the training and inference tractable. HMMs that use this logic is called  discriminant Markov model \citep{ConditionalHMM} or sometimes conditional HMMs in the literature \citep{CHMM}.
In discriminant HMMs, we have a HMM to model each label of the target sequence. Therefore, the model for the sequence of labels is constructed by concatenating the HMM of each label that would become HMM again. The way that neural networks are used in these models is similar to generative HMMs. The probability of the target sequence given the input can not be computed in a tractable manner. However, a dynamic programming approach can be used to efficiently  decode the target sequence, decoding means computing the target sequence that maximizes the posterior.  

% sequential conditional likelihood based generative models ==>> CTC
CTC is another choice for conditional sequence modeling \citep{CTC2006}.
Similar to HMM approach, CTC is applicable when the output sequence is discrete.
It computes the probability via marginalizing over possible alignments between an input sequence and an output sequence.
Since there is a huge number of alignments, computing this probability is inefficient. CTC approximates it by a dynamic programming approach to reduce computation time complexity and makes it tractable.
In most cases of using the CTC loss function, a RNN is used as a probability estimator in every time-step \citep{CTC_Deep_Speech}.
However, this probability can be computed by other structures like a convolution neural network \citep{CTC_LPRNET}.

% sequential conditional likelihood based generative models ==>> comparison
One of the advantages of HMMs or CTC models in contrast to encoder-decoder models is that they created an accurate alignment of the input sequence.
These models have performed very well in speech recognition tasks. But recently encoder-decoder models have shown promising results and beat HMM models in some datasets \citep{Compare1,Compare2}. To solve the alignment problem and improving the performance, the hybrid of CTC and encoder-decoder methods have been proposed recently \citep{Espnet}.

% sequential non-conditional non-likelihood based generative models ==>> SeqGAN, TimeGAN
One important category of non-conditional non-likelihood based sequential models are sequential versions of Generative Adversarial Networks (GANs).
Recently, various methods have been introduced to provide sequential versions of GANs \citep{SeqGAN,TimeGAN}.
One of them is named SeqGAN \citep{SeqGAN} which uses a RL framework.
In SeqGAN, the generator is treated as an RL agent. The states in each time-step are the generated sequence up to the current time-step, and the action is defined as the observation in the next time-step that will be generated. The generator will receive a reward from the discriminator unit corresponding to the generated sequence by the generator. Its optimization strategy for training the generator unit is based on policy gradient \citep{PolicyGradient}.
When the generator unit is frozen, the discriminator unit is trained to distinguish between the generated fake sequence and the real sequence in the dataset.

TimeGAN is another sequential GAN that uses GAN in a recurrent manner in a latent sequence created by a recurrent auto-encoder. It uses a RNN for the generator of the latent sequence. The discriminator uses a  RNN to encode the whole input sequence (the latent sequence of the real input or the latent sequence of the generator output), and the average pooling of the hidden states is used in discrimination implemented by a feed-forward network. In TimeGAN, GAN and auto-encoder networks are optimized jointly.

% sequential non-conditional non-likelihood based generative models ==>> SeqGAN, TimeGAN
One interesting non-likelihood-based sequential conditional model has been proposed in \citep{StoryGAN} as a story visualization task.
The generator is a RNN that apart from noise vector sequence, it gets the encoded sentence as another input in each time-step and also it gets the encoded information of the whole story as another input for the first time-step.
Moreover, it comprises two discriminator units (image and story discriminator) for assuring the coherency of the generated sequence of images besides the quality of them.
Briefly, the image discriminator evaluates the quality of the generated image in every time-step, while the story discriminator gets the dot product of two vectors (concatenation of the encoded generated images and concatenation of the encoded sentences) as input contained coherence information for measuring the coherence among images.

\subsection{The limitation of RMDNs}
% Literature of RMDN
The expressive power of (Recurrent) MDNs is limited by the number of components in the mixture. Using a large number of components may lead to over-parametrization in high-dimensional problems.  In most applications, diagonal GMMs (GMMs with diagonal covariances) has been used in (Recurrent) MDNs for high-dimensional problems which is a limiting assumption on underlying distribution of the data \citep{VariationalRNN,StochasticRNN}. Since the problem is similar for both cased on MDNs and recurrent version of them, we review the solutions that are given in the literature for each of these cases.
 
Different solutions have been proposed in the literature to overcome the over-parametrization problem of RMDNs. The purpose of most of them is to regularize the model in different ways.
Some methods use noise regularization by injecting noise to the training \citep{RegMDN,CDEwithNN}.
It has been observed this kind of regularization smoothes the cost function and partially helps avoiding overfitting problem.
Several methods have tried to introduce an additive regularization term (usually \(\ell_2\)-norm of network's weights) to the loss function \citep{Primitive,RegMDN,BayesianIterpolation}.
A recent method \citep{Primitive} has added an entropy-based loss function instead of \(\ell_2\)-norm of weights to reduce mode and modal collapse problems. Besides, it has introduced a failure loss in some epochs to gather a set of the failed sample during training to improve training performance and avoid under-fitting.

Structural regularization can be considered as another regularization method for solving the over-parametrization problem of RMDNs. For example in Kernel Mixture Networks, only the weights of of each Gaussian kernel is obtained by a neural network \citep{KMN},
and other parameters are fix relative to training data.
The centers of kernels are found by subsampling from training data, and the scale of them are fix.

Another approach which is similar in spirit to our proposed method is mapping the target variable to a new (latent) space. The latent space can be a lower-dimensional representation of the target that allows using more number of components without fear of overfitting. Also, the latent space can be more regular such that it can be fitted with smaller number of components that can even be restricted (for example having diagonal covariance matrices). In \citep{VariationalRNN}, a Variational Auto-Encoder (VAE) is used in every time-step to map the variable to a latent space, then RMDN is used for fitting the sequence in the latent space.

\section{Preliminaries}
\label{sec:preliminaries}
The preliminaries of the proposed method is introduced in this section.

\subsection{\bf Recurrent Mixture Density Network}
\label{sec:mdn}

As mentioned in \SectionRef{sec:introduction}, RMDN tries to model a sequential data distribution with a GMM in every time-step. Therefore to generate the data in each time-step, it is enough to sample from the GMM density.
Since we have a closed-formed density model in RMDNs, the common objective for estimating the parameters is NLL based on the available \textit{i.i.d} samples of the data, \(\{(x^q, y^q)\}_{q=1}^{Q}\), from \(p(x,y)\), wherein \(Q\) is the number of samples, and \(x^q=x^q_{\le T_{x^q}}= \{x_{t}^{q}\}_{t=1}^{T_{x^q}}\) and \(y^q=y^q_{\le T_{y^q}}=\{y_{t}^{q} \in \mathbb{R}^d \}_{t=1}^{T_{y^q}}\) are the conditional and dependent variables, respectively.

Due to the sequential nature of the dependent variable, it is clear that the estimated model \(\hat{p}(y|x)\) is defined in every time-step as \(\hat{p}(y_{t+1}|y_{\le t},x_{\le T})\), and total density is constructed by multiplying them.
The overall structure of RMDN is similar to % \FigRef{fig:proposed_method},
\FigRef{fig:paper}, but with the difference that it does not need to pass the next dependent variable through a NF.
Besides, the covariance matrix does not need any special decomposition like the proposed method.
More details about the training and the sampling of this model are provided below.

\subsubsection{Training}

Given a sequence of
\((x_{\le T_{x}},y_{\le T_{y}})\), RMDN computes the conditional distribution at time \(t+1\) with a mixture model given by
\begin{equation}
	\label{eq:ConditionalDensity}
	p({y_{t+1}}|{y_{\le t}, x_{\le T_{x}}}) =
	\sum\limits_{k = 1}^K {{\alpha _k}({y_{ \le t}, x_{ \le T_{x}}})
	\varphi \left( {{y_{t+1}};{\theta _k}({y_{\le t}}, {x_{ \le T_{x}}})} \right)},
\end{equation}
where, \(p({y_{t+1}}|{y_{\le t}}, {x_{\le T_{x}}})\) is the conditional probability of the sequence at time \(t+1\), \(K\) is the number of components of the mixture, \({\alpha _k}\) is the coefficient of the \({k^{th}}\) component which also is a function of the conditional and the previous dependent variables,
\(\varphi \left( {{y_{t+1}};{\theta _k}({y_{\le t}}, {x_{ \le T_{x}}})} \right)\) is a parametric distribution for \({k^{th}}\) component
that its parameters (\({\theta _k}({y_{\le t}}, {x_{ \le T_{x}}})\))
are functions of the conditional and the previous dependent variables.
GMM is a common choice of a mixture model in the literature of RMDNs.
In case of GMM, the parametric distribution in \EquationRef{eq:ConditionalDensity}, \(\varphi \left( {{y_{t+1}};{\theta _k}({y_{\le t}}, {x_{ \le T_{x}}})} \right)\), is written as the following equation.
\begin{equation}
	\label{eq:Gaussian}
		\begin{split}
		&\varphi \left( {{y_{t+1}};{\theta _k}({y_{\le t}}, {x_{ \le T_{x}}})} \right) = 
		{(2\pi)}^{\frac{-d}{2}}
		{\left|\Sigma_k({y_{\le t}}, {x_{ \le T_{x}}})\right|}^{\frac{-1}{2}}\\
		&\qquad\exp \biggl( \frac{{ - 1}}{2}{{\left( {{y_{t+1}} - \mu ({y_{\le t}}, {x_{ \le T_{x}}})} \right)}}{\Sigma_k ^{ - 1}}
		({y_{\le t}}, {x_{ \le T_{x}}})
		\left( {{y_{t+1}} - \mu ({y_{\le t}}, {x_{ \le T_{x}}})} \right)^T \biggr),
		\end{split}
\end{equation}
where, \(d\) is the dimension of the problem. Also, \({\Sigma _k}\) and \({\mu _k}\) are the covariance matrix and the mean vector of the \({k^{th}}\) component, respectively.

Given the training data \(\left\{ (x^q_{\le T_{x^q}}, y^q_{\le T_{y^q}}) \right\}_{q = 1}^Q\), the equation below describes the criterion of RMDNs in terms of NLL, and it is optimized with gradient descent methods.
\begin{equation}
	\label{eq:SeqNLL}
	\text{NLL}(\mathcal{Y}|\mathcal{X}) =  - \sum\limits_{q = 1}^Q {\sum\limits_{t = 1}^{{T_q}} \log {p(y_{t+1}^q|{y_{\le t}^q}, {x_{ \le T_{x^q}}^q})} }.
\end{equation}

In case of GMM, there is a set of parameters, \(\left\{ {{\alpha _k},{\mu _k},{\Sigma _k}} \right\}_{k = 1}^K\), that are the outputs of a neural network in every time-step.
Since these parameters must satisfy some constraints to be valid parameters, thus the last output of the neural network in every time-step, \(o_t\) and briefly in form of \(o\), is divided into three parts and each part has its own specific activation function to satisfy these constraints.
More details about these parameters are as follows:

\paragraph{Coefficient}$ $
% \textbf{Coefficient:}

In order to achieve the conditional probability equal to one in \EquationRef{eq:ConditionalDensity},
the coefficients must satisfy
\begin{equation}
	\label{eq:Alpha}
	\begin{array}{*{20}{c}}{\sum\limits_{k = 1}^K {{\alpha _k} = 1} ,}&{0 \le {\alpha _k} \le 1}\end{array}.
\end{equation}

Based on the mentioned constraint, the \text{softmax} activation function in \EquationRef{eq:Softmax} is applied on the part of the last layer of the neural network corresponding to coefficients (\({o^\alpha }\) with \(\left| {{o^\alpha }} \right| = K\)) by
\begin{equation}
	\label{eq:Softmax}
	{\alpha _k} = \frac{{\exp (o_k^\alpha )}}{{\sum\limits_{k' = 1}^K {\exp (o_{k'}^\alpha )} }}.
\end{equation}

\paragraph{Mean vector}$ $
% \textbf{Mean vector:}

The second part of the parameters is mean vectors that indicate the position of components in the space.
It does not need to apply any activation function on corresponding outputs to the mean vectors (\({o^\mu }\) with \(\left| {{o^\mu }} \right| = K \times d\)). Therefore, the mean vectors are given by
\begin{equation}
	\label{eq:Mean}
	\begin{array}{*{20}{c}}{{{({\mu _k})}_i} = {{(o_k^\mu )}_i},}&{i \in \left\{ {1,...,d} \right\}}\end{array}.
\end{equation}

\paragraph{Covariance matrix}$ $
% \textbf{Covariance matrix:}

The mean vectors adjust the position of components, while the covariance matrices model the variations of them in different dimensions.
Considering a full covariance matrix involves a hard optimization problem, besides satisfying the positive definite condition for them.
Furthermore, the number of parameters of the covariance matrix, \(K \times \frac{{d(d + 1)}}{2}\), is high for high-dimensional data, 
and directly estimating the whole parameters may overfit the data.
As a consequence, a full covariance matrix has been rarely used in the literature of RMDN.

In most cases for simplicity, the correlation among different dimensions is ignored, and a diagonal covariance matrix is used.
So, it is enough to include the diagonal elements of the matrix be greater than zero.
Therefore, \EquationRef{eq:Variance} is applied on the corresponding parts of the outputs to the covariance matrix (\({o^\Sigma }\) with \(\left| {{o^\Sigma }} \right| = K \times d\)) as
\begin{equation}
	\label{eq:Variance}
	\begin{array}{*{20}{c}}{{{({\sigma _k})}_i} = f {{\left( {o_k^\Sigma } \right)}_i},}&{i \in \left\{ {1,...,d} \right\}},\end{array}
\end{equation}
where, \({\sigma _k}\) contains elements of the diagonal covariance matrix corresponds to \({k^{th}}\) component. Also, \(f\) is a function with a positive range. Two commonly used functions for \(f\) in the literature are \(\exp\) and softplus.

\subsection{\bf Normalizing Flow}
\label{sec:nf}
The main goal of NFs is transferring a simple distribution (commonly a zero-mean spherical Gaussian distribution) to a goal distribution by applying a sequence of invertible transformations.
The overall concept of this network with its notations is depicted in \FigRef{fig:nf}.
\begin{figure}[h]
	\centering
	\includegraphics[width=3.5in]{\FigPath{FlowIdea1.png}}
	\caption{A graphical illustration of NFs has been depicted in this figure. The left is a zero-mean spherical Gaussian distribution; the right is a more complicated distribution.
	}
	\captionsetup{justification=centering,margin=2cm}
	\label{fig:nf}
\end{figure}
By considering a latent random variable \({z_0} \in {\mathbb{R}^d}\) with density function \({p_0}\), NF applies a sequence of invertible transformations to find a new random variable \({z_N} = {f_N^{-1}} \circ ... \circ {f_1^{-1}}({z_0})\), \({z_N} \in {\mathbb{R}^d}\), with density function \({p_N}\).
These combination of invertible transformations is also invertible.
Therefore, its inverse is expressed as \({z_0} = {f_1} \circ ... \circ {f_N}({z_N})\), \({z_0} \in {\mathbb{R}^d}\).
The density function of the latent random variable (\({z_0}\)) is calculated by \EquationRef{eq:NF} based on the change of variable rule.
\begin{equation}
	\label{eq:NF}
	{p_0}({z_0}) = {p_N}({z_N})
	{
		\prod\limits_{n = 1}^N {{{ {\det \left| \frac{{\partial {f_n^{-1}}({z_{n - 1}})}}{{\partial {z_{n - 1}}}} \right| } }}},
	}
\end{equation}
where, \(\det \left| \frac{{\partial {f_n^{-1}}({z_{n - 1}})}}{{\partial {z_{n - 1}}}} \right|\) is the Jacobian determinant of the \(n^{th}\) transformation.
There are different choices for intermediate transformations (\(\left\{ {f_n} \right\}_{n = 1}^N\)) in the literature of NFs \citep{SurveyNF}.
To facilitate the calculations of \EquationRef{eq:NF}, 
the Jacobian determinants of
\(\left\{ {f_n} \right\}_{n = 1}^N\)
have to be efficiently computed.

Apart from simple linear and point-wise non-linear transformation, an important transformation called affine coupling layer has been used in the literature \citep{SurveyNF}.
\label{sec:realnvp}
This transformation can be written as
\(z_{n} = b \odot z_{n-1} + \Big( (1-b) \odot z_{n-1}) \odot \exp (s(b \odot z_{n-1})) + t(b \odot z_{n-1}) \Big)\) where \(b\) is a binary mask operator to partition the input.
The first part of input does not change using this transformation, that is \({z_{n, 1:d'}} = {z_{n-1, 1:d'}}\). 
The rest of it is changed using this transformation by \({z_{n, d' + 1:d}} = {z_{n-1, d' + 1:d}} \odot \exp (s({z_{n-1, 1:d'}})) + t({z_{n-1, 1:d'}})\),
based on a scale and transition function conditioned on the first part.
Normally, the transformation is applied in an alternative pattern throughout the network by swapping two parts in compare to the previous layer \citep{RealNVP}. 

Affine coupling layer is invertible and its inverse is calculated by
\(z_{n-1} = \Big( (1-b) \odot z_{n} - t(b \odot z_{n}) \Big) \odot \exp (-s(b \odot z_{n}))\).
The invertibility does not depend on the invertibility of \(s\) and \(t\) anymore. So, \(s\) and \(t\) can be complex as desired; a deep neural network is usually used for these two units.
Affine coupling layer has the following block-triangular Jacobian matrix
\begin{equation}
\label{eq:Jacobian}
{
	J = \left[ {\begin{array}{ccccccccccccccc}
		{{I_{d'}}}&{{0_{d' \times (d - d')}}}\\
		{{{(\frac{{\partial {z_{n, d' + 1:d}}}}{{\partial {z_{n-1, 1:d'}}}})}_{d'}}}&{\text{diag}{{(\exp (s({z_{n-1, 1:d'}})))}_{d' \times (d - d')}}}
		\end{array}} \right].
}
\end{equation}
The logarithm determinant of Jacobian is simply equal to \( \sum\limits_{i = 1}^{d-d'} s(z_{n-1, 1:d'})_i \),
which does not need the computation of the Jacobian determinant of \(s\) and \(t\).

\section{Proposed method}
\label{sec:proposed}
In this section, we explain the details of our proposed extension to RMDNs named FRMDN that increases the expressive power of RMDNs. Furthermore, an effective decomposition of precision matrices is proposed that can be used in both RMDNs and FRMDNs.

Based on available \textit{i.i.d} samples of the data \(\{(x^q, y^q)\}_{q=1}^{Q}\),  wherein \(x^q=x^q_{\le T_{x^q}}= \{x_{t}^{q}\}_{t=1}^{T_{x^q}}\)  and \(y^q=y^q_{\le T_{y^q}}=\{y_{t}^{q}\}_{t=1}^{T_{y^q}}\) are the conditional and dependent variables,
RMDN models the probability of the dependent variable at each time-step conditioned on depend variables in the previous time-steps and conditional variables \(p(y_{t + 1}|y_{\le t}, x_{ \le T_x})\). The density is given by a GMM which its parameters are the output of a neural network, in every time-step.

To increase the expressive power of RMDN, FRMDN transforms the dependent variable to the new space using a NF and then models the conditional density of transformed variable similar to RMDN. Mathematically speaking, the conditional density
\(p(f(y_{t + 1})|y_{\le t}, x_{ \le T_x})\)
is written as the following equation
\begin{equation}
\label{eq:NLLProposedGMM}
p(f(y_{t + 1})|y_{\le t}, x_{ \le T_x})= \sum\limits_{k = 1}^K \alpha _k(y_{\le t}, x_{ \le T_x})
\varphi (f(y_{t+1};\theta_{\text{flow}});{\theta _k}(y_{\le t}, x_{ \le T_x})),
\end{equation}
where, \(K\) is the number of components, \({\alpha _k}({y_{\le t}, x_{ \le T_x}})\) and \({\theta _k}({y_{\le t}, x_{ \le T_x}})\) are the parameters of the \({k^{th}}\) component, and
\(\varphi \left( {f({y_{t+1}, {\theta_{\text{flow}}}});{\theta _k}({y_{\le t}, x_{ \le T_x}})} \right)\)
is a Gaussian distribution given by \EquationRef{eq:Gaussian}.

We can use \EquationRef{eq:NF} to compute the conditional density of  the dependent variable before transformation by NF:
\begin{equation}
\label{eq:NLLProposedGMMwithFlow}
p(y_{t + 1}|y_{\le t}, x_{ \le T_x}) = p(f(y_{t + 1})|y_{\le t}, x_{ \le T_x})
\prod\limits_{n = 1}^N {{{\left| {\det \frac{{\partial {f_n^{-1}}({z_{n - 1}})}}{{\partial {z_{n - 1}}}}} \right|}
}},
\end{equation}
where, \(f(y_{t + 1})\) is assumed the transformed data, \(z_0=y_{t+1}\), \(z_N=f(y_{t+1})\), \(\left\{ {{z_n} = f_{n}\left( {{z_{n - 1}}} \right)} \right\}_{n = 1}^N\) are the intermediates transformed data by NF, and \(\det \left| \frac{{\partial f_{n}^{-1}({z_{n - 1}})}}{{\partial {z_{n - 1}}}} \right| \) is the Jacobian determinant of the transformation.

A significant amount of parameters for a Gaussian distribution are related to a full precision matrix in \EquationRef{eq:Gaussian}. 
It can become more severe in the case of GMMs, where we have lots of Gaussian densities.
Considering a proper decomposition for
the precision matrix, \({\Sigma _k^{-1}}({y_{\le t}, x_{ \le T_x}}) = D_k({y_{\le t}, x_{ \le T_x}}) + U_k({y_{\le t}, x_{ \le T_x}}){U_k^T({y_{\le t}, x_{ \le T_x}})}\), where every component in the mixture has its specific diagonal matrix (\({D_k}\)) and a non-square matrix \(U_k\),
reduces the number of parameters of the model.
The number of parameters assigned to the full precision matrix in the GMM is \(\mathcal{O}(Kd^2)\), while the proposed decomposition is about \(\mathcal{O}(Kd(1+d'))\). As mentioned before, $d' \ll d$ leads to a linear order relative to the dimension of the problem as \(\mathcal{O}(Kd)\).
It is important to note that the proposed model has been evaluated in two cases, whether the matrix is the output of the neural network (\(U\) is a function of independent variables) or not. We observed that the case where \(U\) is a function of independent variables gives significantly better results, and therefore we only give the results of this case in the experiments.

The graphical view and the pseudo-code of the proposed method are presented in \FigRef{fig:paper} and \AlgoRef{algo:algorithm}, respectively.
Briefly, for generating samples from FRMDN, 
it is enough to follow \(y_{t + 1}^q = g(z_{t + 1}^q) = {f^{ - 1}}(z_{t + 1}^q)\).
The simplicity of this process is related to NF's properties.
It is enough to generate a sample from the corresponding GMM
at every time-step, and then apply an inverse transformation (like \(f^{-1} = {f_N}^{-1} \circ ... \circ {f_1}^{-1}\) or \(g = {g_N} \circ ... \circ {g_1}\)) on it.

\begin{algorithm}
	\caption{Pseudo-code for training and generating samples of FRMDN at time \(t+1\) for \(q^{th}\) data.}
	\label{algo:algorithm}
	\begin{algorithmic}[1]
		\Procedure{\underline{Training}}{}
		\State $\textit{\(x_{\le T_x^q}^q\)},\textit{\(y_{\le t}^q\)} \gets \text{conditional variables at time \(t+1\)}$
		\State $\textit{\(y_{t+1}^q\)} \gets \text{dependent variables at time \(t+1\)}$
		\State $\textit{\(z_{t+1}^q\)} \gets f\text{(\(y_{t+1}^q\))} \text{, where } f \text{ is a NF}$	
		\State $\textit{\(\left\{ {{\alpha _k},{\mu _k},{D_k}, {U_k}} \right\}_{k = 1}^K\)} \gets \text{RNN(\(x_{\le T_x^q}^q, y_{\le t}^q\))}$
		\State $\text{calculate NLL based on \(\left\{ {{\alpha _k},{\mu _k},{D_k}, U_k} \right\}_{k = 1}^K\)}$
		\State $\text{update parameters \(\theta=\left\{\theta_{\text{flow}}, \theta_{\text{gmm}}\right\}\)}$
		\EndProcedure
		\\
		\Procedure{\underline{Generating}}{}
		\State $\textit{\(x_{\le T_x^q}^q\)},\textit{\(y_{\le t}^q\)} \gets \text{conditional variables at time \(t+1\)}$
		\State $\textit{\(\left\{ {{\alpha _k},{\mu _k},{D_k}, {U_k}} \right\}_{k = 1}^K\)} \gets \text{RNN(\(x_{\le T_x^q}^q, y_{\le t}^q\))}$
		\State $\textit{\(y_{t + 1}^q = g(z_{t + 1}^q) = {f^{ - 1}}(z_{t + 1}^q)\)}$
		\State \text{return \(y_{t+1}^q\)}
		\EndProcedure
	\end{algorithmic}
\end{algorithm}

\begin{figure*}[h]
	\centering
	\includegraphics[width=6in]{\FigPath{MainIdea_detail_v2.png}}
	\caption{
		This diagram is a graphical view of the proposed method. Briefly, it uses a RNN to parametrize a GMM by considering a custom decomposition (\(D+UU^T\)) for the precision matrices.
		As mentioned in the text, the matrix \(U\) can be either network output or separate.
		The mentioned GMM is used as a prior distribution in every time-step of a NF.
		}
	\label{fig:paper}
\end{figure*}

\subsection{Efficient Computation of NLL}

Similar to RMDN, the loss function used for training FRMDN is NLL. 
By considering the mentioned idea about applying NFs in \EquationRef{eq:NLLProposedGMMwithFlow} and the considered decomposition, NLL is given by
\begin{align}
\label{eq:NLLGMM}
\color{blue}
\begin{split}
&L\left( {\theta  = \left\{ {{\theta _\text{gmm}},{\theta _\text{flow}}} \right\}} \right)
%{} = {}
=
\\
&%\quad\quad
- \sum\limits_{q = 1}^Q \sum\limits_{t = 1}^{T_q}
\Biggl(
\log \sum\limits_{k = 1}^K
\exp \biggl(
{\log \alpha _k}({y_{\le t}^q, x_{ \le T_x}^q} )
+
0.5
\Bigl(
{-d \log \left( {2\pi } \right)}
\\
% \!\begin{aligned}[t]
&\quad\quad\quad\quad\quad\quad\quad\quad
+
{\log \left|D_k( {y_{\le t}^q, x_{ \le T_x}^q}) + U_k( {y_{\le t}^q, x_{ \le T_x}^q})U_k^T( {y_{\le t}^q, x_{ \le T_x}^q})\right|}
\\
&\quad\quad\quad\quad\quad\quad\quad\quad
-
\left(\beta_{t+1}^q
\left(D_k({y_{\le t}^q, x_{ \le T_x}^q})
%+ U_k({y_{\le t}^q, x_{ \le T_x}^q})U_k({y_{\le t}^q, x_{ \le T_x}^q})^T\right)
+ U_kU_k^T\right)
\beta_{t+1}^{q,T}\right)
\Bigr)
\biggr)
\\
&\quad\quad\quad\quad\quad
+
\sum\limits_{n = 1}^N
\log
{{{ {
\det
\left|
\frac{{\partial {f_n^{-1}}({z_{n - 1, t + 1}^q})}}
{{\partial {z_{n - 1, t + 1}^q}}}
\right|
} }}}
\Biggr),
% \end{aligned}
\end{split}
\end{align}
where, $\beta_{t+1}^q = f(y_{t + 1}^q) - \mu _k^T({y_{\le t}^q, x_{ \le T_x}^q})$. As it can be seen in \EquationRef{eq:NLLGMM}, there are lots of vector and matrix computations.
To facilitate calculations, we follow the following proposition.
% ref: page 420 theorem 18.1.1
\begin{proposition}
	Suppose \(A\), \(B\), and \(C\) are \(d \times d\) invertible, \(d \times d'\), \(d' \times d\) matrices, respectively; then
	\(
	\det(A+BC) = \det(A)\det(I+CA^{-1}B)
	\)
	\citep{Proposition}.
	\label{proposition0}
\end{proposition}

By considering above matrix determinant lemma and a little algebra manipulation, NLL is rewritten as
\begin{equation}
\label{eq:NLLGMM1}
\begin{split}
&L\left( {\theta  = \left\{ {{\theta _\text{gmm}},{\theta _\text{flow}}} \right\}} \right)
=
\\
&\quad\quad
- \sum\limits_{q = 1}^Q \sum\limits_{t = 1}^{T_q}
\Bigg(
\log \sum\limits_{k = 1}^K
\exp \biggl(
{\log \alpha _k}({y_{\le t}^q, x_{ \le T_x}^q})
+
0.5
\Big(
{-d \log \left( {2\pi } \right)}
\\
&\quad\quad
+
% \Big \langle \log\vec{D_k}( {y_{\le t}^q, x_{ \le T_x}^q}), \vec{2} \Big \rangle
% +
2 \log \big( \diag ({D_k}({y_{\le t}^q, x_{ \le T_x}^q})) \big)
+ 2
\log \Big | I_{d'\times d'} + \Omega_{t+1,k}^{q} \otimes \Omega_{t+1,k}^{q, T} \Big |
\\
&\quad\quad
-
\| \Gamma_{t+1,k}^{q} \|_2^2
-
\| B_{t+1,k}^{q} \|_2^2
\Big)
\biggr)
+
\sum\limits_{n = 1}^N
\log
{{{
{
\det
\left|
\frac{{\partial {f_n^{-1}}({z_{n-1, t+1}^q})}}{{\partial {z_{n-1, t+1}^q}}}
\right|
}
}}}
\Bigg),
\end{split}
\end{equation}
{
where
% \( \langle .,. \rangle \) is the dot product operation;
% \(\vec .\) is a vector containing the diagonal elements of a matrix;
\(\diag\) stands for a vector of the diagonal elements of a matrix, and
\(\|.\|_2\) means \(\ell_2\) norm of a vector.
\(\Omega_{t+1,k}^{q}\), \(\Gamma_{t+1,k}^{q}\), \(B_{t+1,k}^{q}\), and \(\beta_{t+1}^q\) are equal to
\(
U^T
\otimes
% \vec D _k^{\frac{-1}{2}}({y_{\le t}^q, x_{ \le T_x}^q})
D _k^{\frac{-1}{2}}({y_{\le t}^q, x_{ \le T_x}^q})
\),
\(
\beta_{t+1}^q
\otimes
{U_k}({y_{\le t}^q, x_{ \le T_x}^q})
\),
\(
\beta_{t+1}^q
\odot
% \vec D _k^{\frac{1}{2}}({y_{\le t}^q, x_{ \le T_x}^q})
D _k^{\frac{1}{2}}({y_{\le t}^q, x_{ \le T_x}^q})
\), and
\(f(y_{t + 1}^q) - \mu _k^T({y_{\le t}^q, x_{ \le T_x}^q})\), respectively, with a tensor-multiplication operator \(\otimes\),
and an element-wise multiplication \(\odot\).
}

\section{Experiment}
\label{sec:experiment}

In this section more details about the proposed method in comparison with other approaches are described.
The experimental description of the three applications used in this paper is given in \SectionRef{sec:ExpSetting}.
The results of each application are presented separately in \SectionRef{sec:results}.

\subsection{Experimental Settings}
\label{sec:ExpSetting}
The settings of three experiments used in this paper, namely image sequence modeling, speech modeling, and single image modeling, are presented in this part. All experiments are done by choosing different optimizer methods (Adam, RMSProp) and different activation functions for the diagonal elements of the covariance \(\exp , \text{softplus}\) several times. The setting resulting in the best performance is given in this section. To tackle the numerical instability for the first application, the diagonal elements of covariances are clipped outside the range \([0.1, 1.9]\).

\paragraph{Application 1: image sequence modeling}$ $

\label{sec:world}

In this paper, we model an image sequence in the spirit used in a successful model-based reinforcement learning framework called the world model \citep{WorldModelNips}.
This cognitive-based framework consists of three major parts: vision model so as to understand the environment, memory model in order to save events and predict the future, and finally the controller to make an accurate decision based on the two previous parts to react in the environment.
In this paper, we modify the original world model \citep{WorldModelNips} by replacing the RMDN of the memory unit with FRMDN to enrich the probability model.
In this paper, we examine our model on two environments,
{\href{https://gym.openai.com/envs/CarRacing-v0}{\color{black} Car-Racing}} (the same as the original paper \citep{WorldModelNips}) and {\href{https://pypi.org/project/gym-super-mario-bros}{\color{black} Super-Mario}}.

The visual unit provides a compact representation of the input using a VAE. Accordingly, each observation is transferred to a latent vector.
This unit uses a convolutional VAE to encode each frame to a latent variable. 
Firstly, each colored image is resized to a \(64\times64\times3\) image and is passed to a four layers convolution neural network with a stride of 2 to obtain the mean and covariance parameters of a Gaussian density used to model the latent variable.
The dimensionality of the latent vector is \(32\) in our experiments.
The decoder is a deconvolution network having a ReLU activation function in all layers except the last layer that has a Sigmoid activation function to force the outputs be between 0 and 1. The cost function is the combination of the \(\ell_2\) distance and the Kullback-Leibler loss.
We experiment with a different setting, optimum results are gained by Adam optimizer, the learning rate is set to 0.001, and batch size is equal to 128.

In \citep{WorldModelNips}, the memory unit tries to estimate the probability distribution of the latent variable in the next time-step based on the values of the previous time-steps and the action in the current time-step, that is \(p({z_{t + 1}}|{z_{ \le t}},a_t)\) using a RMDN. 
The memory unit applies a diagonal GMM for estimating the mentioned conditional distribution, for which its parameters are estimated by a RNN.
\citep{WorldModelNips} uses a RNN (one layer LSTM with 256 hidden units) combined with a GMM layer at the last layer to model \(p(z_{t+1})\). It uses a diagonal GMM with 5 components and the model is trained for 20 epochs with the RMSProp optimizer.

As mentioned earlier, we replace the RMDN with FRMDN in this paper.
Our model consists of two parts namely RMDN and NF, and for a fair comparison we  fix the first part (except the GMM layer) and only adjust NF and GMM. The NF consist of two affine coupling block explained in \SectionRef{sec:nf}.
The function \(s\) used in the NF is chosen to be a two layer neural network with a \text{LeakyReLU} activation function for the first layer and a \text{Tanh} activation function for the last layer.
The function \(t\) is also similar to \(s\), except it does not have the last activation function.
It is worth mentioning that the structure has been adjusted by trial and error based on common models for blocks of NFs.

\paragraph{Application 2: speech modeling}$ $
\label{sec:speech}

As the second application, we employ FRMDN for speech modeling.
In this application, every raw audio signal is considered to be a sequence of \(T\) consecutive frames.
Each frame contains 200 samples of audio signals.
Unlike conventional audio processing models, no feature extraction method is used.
The main reason is that we want to be comparable with Variational RNN (VRNN) \citep{VariationalRNN}, which models waveform of speech directly without feature extraction.
The most important reason for our comparison with VRNN is that this article also improves the performance of RMDN by introducing latent variables and using variational inference approach, while we try to employ NF to improve it.
We experiment on three datasets named Blizzard, TIMIT, and Accent same as VRNN \citep{VariationalRNN}.
Normalizing by the global mean and standard deviation of the training data is the employed pre-processing method for all datasets.
Also, as in \citep{VariationalRNN}, Accent and Blizzard are shortened to 0.5 seconds.

We use exactly the same structure for the RNN part of our FRMDN as in \citep{VariationalRNN}.
For the second partition, as mentioned in \SectionRef{sec:nf}, our choices for \(s\) and \(t\) are four consecutive linear layers followed by \text{LeakyReLU} activation functions embedded in two affine coupling blocks NF.

All in all, a description of the structural details specific to each data is presented as follows.

\begin{itemize}
	\item {\bf Blizzard}
	\citep{Blizzard}
	 contains 147248 audio {\it .wav} files, which we use almost half of it (63583 {\it .wav} files).
	We randomly separate 90 percent of the data as the train data  (57223 {\it .wav} files). The rest of the data is evenly split to test and validation sets (3180 {\it .wav} files).
	A sampling frequency of 16 kHz is considered for this data because of being compared with the results of VRNN.
	An Adam optimization, a learning rate of 0.0003, a mini-batch of size 128, and 4000 hidden number is used in RNN part.
	\item {\bf TIMIT}
	 is an acoustic-phonetic speech corpus, while we only use its raw audio files directly.
	The overall setting that is used here is Adam optimizer, a learning rate of 0.001, a min-batch of 64, and 2000 hidden numbers in RNN.
	This dataset has its own data partitioning to train and test data. Each segment has about 1680 {\it .wav} files.
	\item {\bf Accent}
	\citep{Accent}
	 contains 2921 audio samples of different nationalities of speakers reading an English paragraph.
	We randomly separate 90 percent of the data (about 2628 {\it .wav} files) as training data, and the rest is equally divided to test and validation data (each of them is about 146).
	Besides, a recurrent neural network with 2000 hidden units, an Adam optimization method with a learning rate of 0.001, a mini-batch of size 128, and a sampling frequency equal to 16 kHz are selected for this dataset to have a fair comparison with VRNN.
\end{itemize}
\paragraph{Application 3: single image modeling} $ $
\label{sec:image}

Due to the success of RMDN in density estimation, we apply the proposed method to find a class conditional probability distribution \(p(x|y) = \prod \limits_{t=1}^{t=T_x} p(x_{t}|x_{< t}, y) \), where \(x\) is an image and \(y\) is its true label.
So the distribution of every image is found by \(p(x)=\sum \limits_{y} p(x|y)p(y)\), where \(y\) is a one-hot vector of length class number, and \(p(y)\) is the uniform distribution over \(y\), i.e. \(p(y)=\frac{1}{C}\) with \(C\) class number.

As the final application in this paper, we employ two following image datasets for conditional single image modeling:

\begin{itemize}
	
	\item {\bf MNIST}
	\citep{mnist}
	includes 70k one channel handwritten digits images with dimensionality 28$\times$28.
	It contains 50k, 10k, and 10k samples as train, test, and validation sets, respectively.
	The class number is equal to the number of digits, i.e. 10.
	
	\item {\bf CIFAR10}
	\citep{cifar10}
	contains
	105k natural images with dimensionality 32$\times$32$\times$3.
	It contains 90k, 10k, and 5k samples as train, test, and validation stes, respectively.
	All images are categorized into 10 classes.
	So, \(y\) is a one-hot vector of length 10 for every class.
	
\end{itemize}

To be comparable with some state-of-the-art methods \citep{MAF, RealNVP}, the major pre-process methods are the same as  \citep{RealNVP}.
After adding uniform noise to the data, they are transferred to a new space by \(z=\sigma(\lambda + (1-2\lambda)x)\), where \(\sigma\) is the \text{sigmoid} function, and \(\lambda\) is equal to \(1e^{-6}\) for MNIST and \(5e^{-2}\) for CIFAR10.
In the case of CIFAR10, a horizontal flip augmentation is applied on the training data.

Generating image data conditioned on the label is shown in \FigRef{fig:DensityEstiation}. Each image is considered as a sequence, where each row is seen as an element in this sequence. 
In this figure, the purple and green dash lines are the direction of transferring samples in RMDN and FRMDN, respectively.

\begin{figure}[t]
	\centering
	\includegraphics[width=6in]{\FigPath{DensityEstimation.png}}
	\caption{
	The process of generating an image from the proposed models conditioned on a label, when the image is defined as a sequence.
	At each time-step, a row of the image is generated.
	In RMDN, each row is generated by directly sampling from the GMM (purple dashed line).
	In FRMDN, the sample generated by GMM is also passed through the inverse of the NF (green dashed line).}
	\label{fig:DensityEstiation}
\end{figure}

The architecture of the proposed method in this experiment is comprised as follows.
At first, the input is passed to a RNN with 32 hidden units followed by a nonlinear layer to map the hidden state to the parameter space of a GMM.
% In the case of MNIST, it outputs a GMM with 10 components and in the case of CIFAR10, it outputs a GMM with 10 components.
It outputs a GMM with 10 components in both cases of MNIST and CIFAR10.
The used NF model in this architecture is similar to the one used in the speech modeling application.
Adam optimizer with 0.001 learning rate and batch size 256 is the general setup for this application.

\subsection{Experimental Results}
\label{sec:results}
Details of the performed experiments\footnote{The implementation of the proposed methods are available in the following link: \url{http://visionlab.ut.ac.ir/resources/frmdn.zip}.} for the three mentioned applications, namely image sequence, speech and image modeling, are provided in this section.

\paragraph{Image sequence modeling}$ $
\label{sec:result_memory}

Based on the details mentioned in \SectionRef{sec:ExpSetting}, one of our experiments is related to a reinforcement learning problem named the world model.
In this section, the experiments related to the memory unit of the world model for two environments (Car-Racing and Super-Mario) are presented.	
The results for the test data are reported in \TableRef{tab:NLL_new}.
In both environments, FRMDN with \(K=5\) reaches the best performance among all configurations. Furthermore for similar structures, the performance of FRMDN is better than RMDN. Increasing the rank of matrix \(U\) (equivalent to greater \(d'\)) gives better results in terms of NLL for both RMDN and FRMDN.

\begin{table}[thb]
	\caption{
	{NLL values of the RMDN and the FRMDN for test data of two environments and the different number of components of the GMM are reported here.}
	The best value for each component in each environment is shown in bold.
	}
	\label{tab:NLL_new}
	\setlength{\tabcolsep}{1.\tabcolsep}
	\centering
	\begin{tabular}{lcccccccc}
		\hline
		{} & \multicolumn{4}{c}{Car-Racing} & \multicolumn{4}{c}{Super-Mario}\\
		Model & K=1 & K=3 & K=5 & K=1 & K=3 & K=5 \\
		
        \hline
        \hline
		
		RMDN (diagonal) &  2.41 &  2.41 & 2.41 & 1.54 & 1.53 & 1.53 \\
		
		FRMDN (diagonal) & 2.41 & 2.41 & 2.41 & 1.49 & 1.49 & 1.49 \\
		
		\hline
		
		RMDN ($d'=1$)
		& 2.41 & 2.41 & 2.39 & 1.48 & 1.45 & 1.45 \\
		
		RMDN ($d'=2$)
		& 2.41 & 2.40 & 2.36 & 1.47 & 1.36 & 1.37 \\
		
		RMDN ($d'=3$)
		& 2.40 & 2.37 & 2.36 & 1.38 & 1.32 & 1.33 \\
		
		\hline
		
		FRMDN($d'=1$)
		& 2.41 & 2.39 & 2.39 & 1.43 & 1.40 & 1.40 \\
		
		FRMDN ($d'=2$)
		& 2.41 & 2.36  & 2.36 & 1.37 & 1.33 & 1.33 \\
		
		FRMDN ($d'=3$)
		& 2.40 & \boldmath \(2.35\) & \boldmath \(2.35\) & 1.36 &  \(1.29\) & \boldmath \(1.28\) \\
		
		\hline
	\end{tabular}
\end{table}

\paragraph{Speech modeling} $ $

\label{sec:result_speech}

As discussed in \SectionRef{sec:speech}, speech modeling is another application in our experiments.
We apply our proposed methods to three speech datasets named Blizzard \citep{Blizzard}, TIMIT, and Accent \citep{Accent}.
Our results are comparable with VRNN \citep{VariationalRNN} and other baseline methods.
The best results of the proposed method in terms of the NLL are reported in \TableRef{tab:speech_new}.
We performed experiments on the different decompositions of the precision matrix (similar to the previous section).
Here, we did not observe any improvement by including a low-rank term in the precision matrix, therefore, we only report the case of using a diagonal precision matrix.
As it can be seen in \TableRef{tab:speech_new}, In all experiments, FRMDN reaches better results in terms of NLL than RMDN. We use the same model as in \citep{VariationalRNN} for the case of RMDN, but as in can be seen in the table, we get different values for NLL. We do not know the reason for this discrepancy. But as in the table, the improvement attained by FRMDN over RMDN is more than the one attained by VRNN over RMDN.

\begin{table}
	\caption{Comparison between the best results 
		of the proposed method (RMDN and FRMDN) and VRNN \citep{VariationalRNN} in terms of NLL for three speech datasets. Diagonal precision matrices are used in RMDN as in \citep{VariationalRNN}, and also the FRMDN.
	}
	\label{tab:speech_new}
	\centering
	\begin{tabular}{lccc}
		\hline
		Model & Bizzard & TIMIT & Accent \\

		\hline
		\hline
		
		RMDN (K=1)\citep{VariationalRNN} & -3539 & 1900 & 1293 \\
		
		RMDN (K=20)\citep{VariationalRNN} & -7413 & -26643 & -3453 \\
		
		VRNN (K=20)\citep{VariationalRNN} & \boldmath \(\le -9170\) & \(\le -28982\) & \boldmath \(\le -4319\)\\
	
		\hline
		RMDN (K=20) & -4272 & -27674 &   -3063 \\		
		FRMDN (K=20) & -7961  & \boldmath \(-32505\)   & \boldmath \(-4319\) \\
		
		\hline
	\end{tabular}
\end{table}

\paragraph{Image modeling}$ $
\label{sec:result_density}

We apply RMDN and FRMDN with different decompositions on the MNIST and the CIFAR10 datasets as a final experiment.
The results are presented in \TableRef{tab:density}.
We compare our results with the results of RealNVP \citep{RealNVP} and some state-of-the-art methods like MADE \citep{MADE}, and MAF \citep{MAF}.
As it can be seen in the table, FRMDN attains better performance in comparison to RMDN and other methods in terms of the NLL. Increasing \(d'\) also leads to increasing the performance for both methods.

\begin{table}[thbp]
	\caption{
		The results for different setting of the proposed method and some state-of-the-art methods like RealNVP \citep{RealNVP}, MADE \citep{MADE}, and MAF \citep{MAF}.
		The results are reported in terms of NLL for two image datasets.
		The reported results in the first segment of this table are based on the best reported values in \citep{MAF}. The number of components (\(K\)) for RMDN and FRMDN is 10.
		{
		The first segment of the table includes the best test NLL error bars (across 5 trials of optimization settings).
        The rest of the table contains the best NLL of test data for the proposed method.
        }
	}
	
	\label{tab:density}
	\centering
	\begin{tabular}{lcc}
        \hline
		Model & MNIST & CIFAR10  \\
		
		\hline
		\hline
		
% 		Gaussian  & -1344.7 $\pm$ 1.8 & 2030 $\pm$  41 \\
% 		MADE (K=1)  & -1361.9 $\pm$  1.9 & 187 $\pm$  20 \\
% 		MADE (K=10)  & \boldmath \(-1030.3 \pm 1.7\) & -119 $\pm$  20 \\
% 		RealNVP (K=1)  & -1326.3 $\pm$ 5.8 & 2642 $\pm$ 26 \\
% 		MAF (K=1)  & -1302.9 $\pm$  1.7 & 2983 $\pm$ 26 \\
% 		MAF (K=10)  & -1092.3 $\pm$  1.7 & 2936 $\pm$ 26 \\
		
% 		\hline
% 		RMDN (diagonal) & -1209.86 & 2649.27 \\		
% 		RMDN ($d'=1$) & -1206.11 & 2811.65 \\
% 		RMDN ($d'=2$) & -1195.41 & 3078.6  \\
% 		RMDN ($d'=3$) & -1189.89 & 3096.23  \\
		
% 		\hline
% 		FRMDN (diagonal) & -128.58 & {\color{black}3032.25}  \\		
% 		FRMDN ($d'=1$) & -189.51 & {\color{black}3135.29} \\
% 		FRMDN ($d'=2$) & -137.29 & 3146.75  \\
% 		FRMDN ($d'=3$) & 89.89 & 3159.8 \\
		
		Gaussian  & 1344.7 $\pm$ 1.8 & -2030 $\pm$  41 \\
		MADE (K=1)  & 1361.9 $\pm$  1.9 & -187 $\pm$  20 \\
		MADE (K=10)  & \(1030.3 \pm 1.7\) & 119 $\pm$  20 \\
		RealNVP (K=1)  & 1326.3 $\pm$ 5.8 & -2642 $\pm$ 26 \\
		MAF (K=1)  & 1302.9 $\pm$  1.7 & -2983 $\pm$ 26 \\
		MAF (K=10)  & 1092.3 $\pm$  1.7 & -2936 $\pm$ 26 \\
		
		\hline
		RMDN (diagonal) & 1209.86 & -2649.27 \\		
		RMDN ($d'=1$) & 1206.11 & -2811.65 \\
		RMDN ($d'=2$) & 1195.41 & -3078.6  \\
		RMDN ($d'=3$) & 1189.89 & -3096.23  \\
		
		\hline
		FRMDN (diagonal) & 196.79 & -3032.25 \\		
		FRMDN ($d'=1$) & 189.51 & -3135.29 \\
		FRMDN ($d'=2$) & 137.29 & -3146.75  \\
		FRMDN ($d'=3$) & -89.89 & -3159.8 \\

		\hline
	\end{tabular}
\end{table}

Finally, some class conditioned generated images for MNIST using FRMDN with diagonal covariances are shown in \FigRef{fig:generate}.
In the case of challenging CIFAR10 dataset, same as similar state-of-the-art methods like MAF \citep{MAF}, the generated images are very blurry and noisy, and we choose to omit  those results in this paper.
% For the case of challenging CIFAR10 dataset, The generated images are a bit blurry and noisy but still clear.
Attaining better NLL does not necessarily mean having better generation as mentioned in the literature \citep{PML}.
As it was seen, the proposed method beats the state-of-the-art methods in terms of NLL, and therefore, it can be used in applications in which having models with better NLL is important.

\begin{figure}
	\centering
	\subfloat[MAF \citep{MAF}]{{\includegraphics[ height=8.55cm]{\FigPath{MAF_mnist.png}} }}
	\subfloat[FRMDN]{{\includegraphics[height=8.5cm]{\FigPath{mnist_update_removed_space.png}} }}
% 	\\
% 	\centering
% 	\subfloat[MAF (CIFAR10) \citep{MAF}]{{\includegraphics[width=6.2cm, height=8cm]{\FigPath{MAF_cifar.png}} }}
% 	\subfloat[FRMDN (CIFAR10)]{{\includegraphics[width=6.2cm, height=8cm]{\FigPath{cifar_update.png}} }}
	\caption{%\color{blue}
% 	Class conditioned generated samples by MAF \citep{MAF} and diagonal FRMDN for MNIST and CIFAR10. Each row corresponds to one label.
Class conditioned generated samples by MAF \citep{MAF} and diagonal FRMDN for MNIST.
	}
	\label{fig:generate}
\end{figure}

\section{Conclusion}
\label{sec:conclusion}
One of the contributions of this paper was using a NF structure in RMDNs.
For the cases where data is not well-clustered, using NF can be useful to transform the data into a well-clustered space where it can be modeled by a GMM. Experimental results for some applications show the favorable performance of FRMDN in comparison to RMDN, which confirms this intuition.
We also consider a decomposition of the precision matrix \(\Sigma^{-1}=D + UU^T\) in GMM, where both the diagonal matrix \(D\), and the low-rank matrix \(U\) can be the output of the RNN in both RMDN and FRMDN.
This idea improves the modeling power of covariance matrices over commonly used diagonal matrices in high-dimensional problems, with a negligible increase in the parameters and computations.

Numerous experiments were performed on three applications, namely image sequence modeling, speech modeling, and single image modeling. The results in all applications showed the superiority of FRMDN to RMDN in terms of NLL. The results also show superior/competitive performance achieved by FRMDN in compare to other state-of-the-art methods. For single image and image sequence modeling, the performance was improved by adding low-rank terms to the precision matrices.
\begin{figure} 
	\begin{comment}
	\centering
% 	\subfloat[a\label{fig:TwoTheta_a}]{
	\includegraphics[width=1.\linewidth]{\FigPath{future_work.png}}
% 	}
% 	\\
% 	\subfloat[b\label{fig:TwoTheta_b}]{
% 	\includegraphics[width=0.4\linewidth]{\FigPath{future_work_block.png}}}
% 	\caption{(a) next work. (b) next bock.}
	\label{fig:TwoTheta} 
	\end{comment}
	\centering
	\includegraphics[width=3.5in]{\FigPath{future_work_new.png}}
	\caption{A possible extension of the proposed method, where the RNN outputs both the parameters of a NF in addition to the GMM.}
	\captionsetup{justification=centering,margin=2cm}
	\label{fig:TwoTheta}
\end{figure}

An interesting extension to the proposed model is shown in \FigRef{fig:TwoTheta}, where the NF can also be parameterized by some outputs of the RNN.
Another interesting direction of a future work will be a successful decomposition like \(\Sigma^{-1}=UDU^T\) \citep{TiedGMM} with a shared square or non-square matrix \(U\) among different precision matrices in the GMM.

%\bibliography{references}

\end{document}